\documentclass[a4paper,conference]{IEEEtran}
\usepackage{xcolor}

\usepackage[noadjust]{cite}
\ifCLASSINFOpdf
  \usepackage[pdftex]{graphicx}
  \DeclareGraphicsExtensions{.pdf,.jpeg,.png}
\else
  \usepackage[dvips]{graphicx}
  \DeclareGraphicsExtensions{.eps}
\fi
\usepackage{amsmath}
\usepackage{physics}

\usepackage{algorithmic}
\usepackage{array}
\ifCLASSOPTIONcompsoc
  \usepackage[caption=false,font=normalsize,labelfont=sf,textfont=sf]{subfig}
\else
  \usepackage[caption=false,font=footnotesize]{subfig}
\fi

\usepackage{stfloats}
\usepackage[]{hyperref} %\usepackage[hidelinks]{hyperref}
\usepackage{multirow}

\begin{document}
\title{CorrLoss: Integrating Co-Occurrence Domain Knowledge for Affect Recognition}
%TODO editor wegen veränderter Überschrift schreiben

%\author{\IEEEauthorblockN{Michael Shell}
%\IEEEauthorblockA{School of Electrical and\\Computer Engineering\\
%Georgia Institute of Technology\\
%Atlanta, Georgia 30332--0250\\
%Email: http://www.michaelshell.org/contact.html}
%\and
%\IEEEauthorblockN{Homer Simpson}
%\IEEEauthorblockA{Twentieth Century Fox\\
%\and
%\IEEEauthorblockN{James Kirk\\ and Montgomery Scott}
%\IEEEauthorblockA{Starfleet Academy\\
%San Francisco, California 96678--2391\\


\author{\IEEEauthorblockN{Ines Rieger\IEEEauthorrefmark{1}\IEEEauthorrefmark{2},
Jaspar Pahl\IEEEauthorrefmark{1}\IEEEauthorrefmark{2},
Bettina Finzel\IEEEauthorrefmark{2} and
Ute Schmid\IEEEauthorrefmark{2}}
\IEEEauthorblockA{\IEEEauthorrefmark{1}Fraunhofer IIS, Fraunhofer Institute for Integrated Circuits IIS, Erlangen}
\IEEEauthorblockA{\IEEEauthorrefmark{2}University of Bamberg, Cognitive Systems Group}
\IEEEauthorblockA{Email: \{ines.rieger, jaspar.pahl, bettina.finzel, ute.schmid\}@uni-bamberg.de}}


\maketitle
%keywords: facial expression, affective computing, human-machine interaction

\begin{abstract}
Neural networks are widely adopted, yet the integration of domain knowledge is still underutilized. We propose to integrate domain knowledge about co-occurring facial movements as a constraint in the loss function to enhance the training of neural networks for affect recognition. As the co-ccurrence patterns tend to be similar across datasets, applying our method can lead to a higher generalizability of models and a lower risk of overfitting. We demonstrate this by showing performance increases in cross-dataset testing for various datasets. We also show the applicability of our method for calibrating neural networks to different facial expressions.
\end{abstract}
\begin{figure*}[!t]
	\centering
	\subfloat[Happy face example from \cite{seuss2019emotion}.
	]{\includegraphics[width=1.5cm]{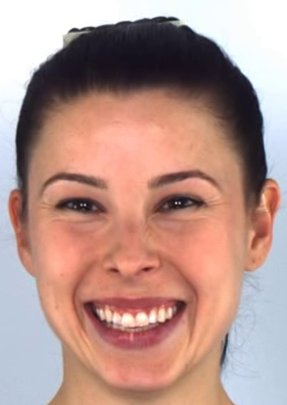}
		\label{fig:happyface}}
	\hfil
	\subfloat[\textbf{Ground truth correlations} of facial AUs for happy faces: There is eye brow movement (AU 1-2), cheek raising and lid tightening (AU 6-7), and mouth movements that extend also to the cheek and chin (AU 12-24).
	]{\includegraphics[width=5cm]{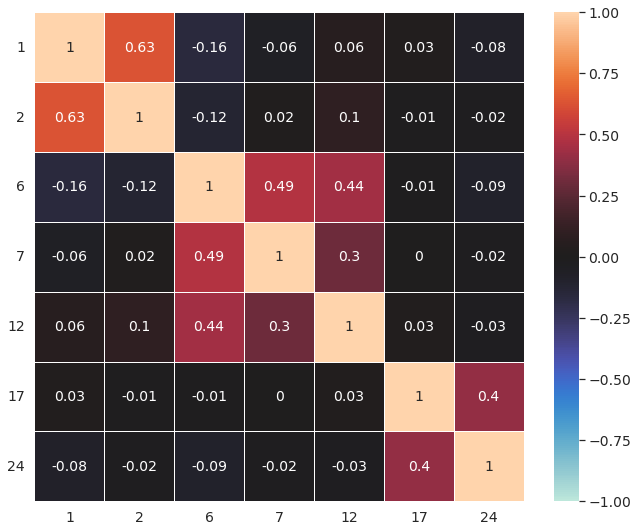}
		\label{fig:t1_pred_gt}}
	\hfil
	\subfloat[These are predictions for happy faces in test data, when trained with \textbf{binary crossentropy}. The neural network learns the correlations automatically to a certain degree.
	]{\includegraphics[width=5cm]{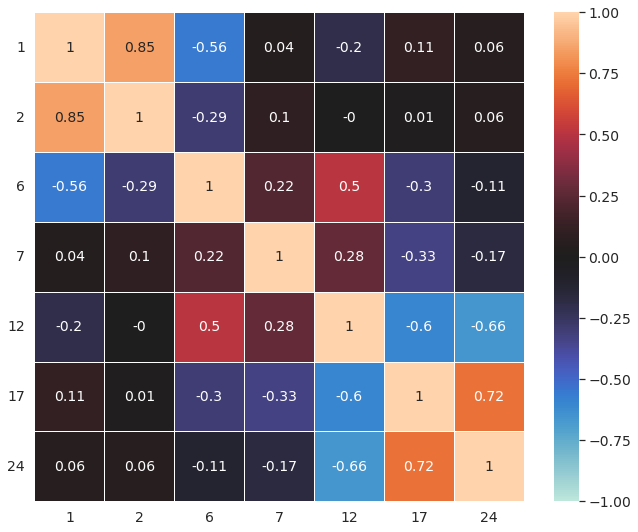}
		\label{fig:t1_pred_bce}}
	\hfil
	\subfloat[These are predictions when trained with \textbf{binary crossentropy (BCE) and CorrLoss}. CorrLoss forces the NN to learn the true correlations between AUs. This matrix is more similar to the ground truth correlation compared to when only trained with BCE.
	]{\includegraphics[width=5cm]{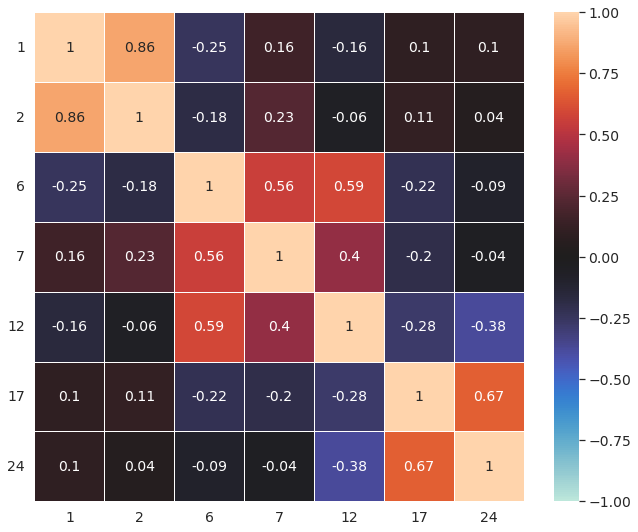}
		\label{fig:t1_pred_corrloss}}
	\caption{Intuition on how well the model learns ground truth correlations with our CorrLoss regularization term. The example stems from Section \ref{calibrationonfacialexpressions}, where we finetune models on facial expressions. %Go there for more details and quantitative evaluation.
	}
	\label{fig:t1_pred}
\end{figure*}

\ifCLASSOPTIONpeerreview
\begin{center} \bfseries EDICS Category: 3-BBND \end{center}
\fi
\IEEEpeerreviewmaketitle
A purely data-driven approach for training neural networks may reach its limits, for example, when there is training data of low quality or when there are constraints the model must satisfy such as natural laws or other regulations~\cite{von2019informed}. Additionally, as neural networks become more complex, the need for interpretability increases. Integration of domain knowledge can tackle all of these disadvantages by forcing the neural network to adhere to constraints, which also enhances the interpretability.
\\
In our approach, we propose to integrate domain knowledge on co-occuring target classes directly in the loss function to enhance affect recognition models. For our experiments, we concentrate on detecting facial movements called Action Units (AUs). AUs are a psychological framework to describe distinct, objective facial muscle movements such as lowering the brow, or raising the cheek in a modular way. %describe facial behavior as they categorize distinct facial muscle movements such as lowering the brow, or raising the cheek. Generally, AUs provide a modular and objective set to describe facial expressions. 
For more information about AUs see the description by Ekman and Friesen~\cite{ekman2002facial} and the survey on automatic facial AU analysis by Zhi et al.~\cite{zhi2020comprehensive}. \\
One disadvantage of affective computing and especially AU datasets are the varying properties regarding their recording conditions, i.e. \textit{in-the-lab} vs. \textit{in-the-wild} or \textit{acted} vs. \textit{natural}. Training on datasets with very specific properties leads to models which suffer from bad generalizability  and therefore do not evaluate well on datasets with different properties~\cite{ertugrul2019} in a cross-dataset setting. %Furthermore, training datasets for AU detection in general are scarce, because the annotation process for AUs takes a considerable amount of time and requires certified expert annotators.\\
%(see for example: Facial Action Coding System (FACS)~\cite{ekman2002facial}, or survey on automatic facial AU analysis~\cite{zhi2020comprehensive}). 
%
%
Domain knowledge can tackle this disadvantage, since it is to a certain degree disentangled from the dataset properties (e.g. recording setting or subject metadata) and therefore provides general information about the task.
%carries potential to counteract label noise and over- or underrepresented classes in the training data as it is to a certain degree disentangled from the dataset properties (e.g. recording setting or subject metadata). 
For AUs, domain knowledge in the form of co-occurrences exist due to the fact that facial expressions such as emotions, pain or stress activate specific subgroups of AUs~\cite{du2014compound}. Furthermore, because of the anatomically predetermined dependence of movements in the face, the contraction of muscles can lead to the activation of several AUs.  Since the patterns for the same facial expression are similar across subjects, we propose to use the co-occurrence information to enhance the model's generalizability and to calibrate models on distinct facial expressions.\\
%enhance the models calibrate Affect Recognition models on facial expressions by using the co-occurrence information.\\
%
%In the field of affective computing and especially AU detection, we face several challenges regarding training datasets. 
%Affective computing datasets have varying properties regarding their recording conditions, i.e. \textit{in-the-lab} vs. \textit{in-the-wild} or \textit{acted} vs. \textit{natural}. 
%Training on datasets with very specific properties leads to models which suffer from bad transfer capabilities and therefore do not evaluate well on datasets with different properties~\cite{ertugrul2019}. Furthermore, training datasets for AU detection in general are scarce, because the annotation process for AUs takes a considerable amount of time and requires certified expert annotators.
%To enhance the model's generalizability, we propose to also use the co-occurrence of facial movements as domain knowledge. This knowledge carries potential to counteract label noise and over- or underrepresented classes in the training data as it is to a certain degree disentangled from the dataset properties (e.g. recording setting or subject metadata). 
%
More specifically, we formulate the co-occurrence information as a weighted regularization term (\textbf{CorrLoss}) to optimize positive and negative AU correlations and combine it with binary crossentropy loss (BCE). In contrast other approaches that model the co-occurrence information in a hypothesis space (see Section~\ref{relatedwork}), we formulate the co-occurrence constraint as a regularization term. We find this a lightweight solution, which is furthermore flexible to steer as the domain knowledge does not need to be modeled first.
%TODO mehr, der corrloss tut was er soll?
For highlighting the interpretability aspect, we provide visualizations of the ground truth and learned co-occurrences that can be inspected with respect to plausibility (Fig.~\ref{fig:t1_pred}). To the best of our knowledge, we are the first to formalize a co-occurrence constraint directly in the loss function and to finetune with this knowledge on facial expressions. We are also the first to conduct a comprehensive cross-dataset evaluation for assessing the generalizability of using co-occurrence knowledge.
%
%The generalization ability of models can be tested in a cross-dataset cross-domain evaluation.\\
%
Concisely, we answer the following research questions. Does CorrLoss improve
\begin{enumerate}
	\item \textbf{within dataset} performance?
	\item \textbf{cross-dataset} performance?
	\item \textbf{calibration} on facial expressions?
\end{enumerate}
%
%TODO erst später?
To evaluate our approach we use several AU benchmark datasets~(Section~\ref{sec:datasetprep}): BP4D~(\cite{zhang2013high,zhang2014bp4d}), CK+~\cite{kanade2000comprehensive,lucey2010extended}, and GFT~\cite{girard2017sayette}, Actor Study~\cite{seuss2019emotion}, AffWild2~(\cite{kollias2017recognition,kollias2018aff,kollias2018multi,kollias2019expression}), and EmotioNet (manually annotated part)~\cite{emotionet}.\\
Our key findings are: (1)~When evaluating the within dataset performance, using our CorrLoss decreases the variance over different data folds, but does not significantly increase the mean results~(Section~\ref{crossdatasetevaluation}). The lower variance over several different data folds can indicate enhanced robustness. We can also observe a decreased risk of overfitting in the training. 
%In general, our trained model reaches competitive performance in comparison with state-of-the-art approaches for the within dataset evaluation~(Section~\ref{stateoftheartcomparison}).
(2)~When evaluating our CorrLoss in a cross-dataset setting, the mean performance increases and variance decreases for most datasets compared to our baseline. This means that \mbox{CorrLoss} can increase the robustness and generalizability of the model~(Section~\ref{crossdatasetevaluation}). This is also reflected in our state-of-the-art comparison as our model outperforms in the cross-dataset evaluation~(Table \ref{tab:bp4dsota}). (3)~We can see a performance gain when we calibrate our trained models with CorrLoss on specific facial expression tasks like happiness or pain~(Section~\ref{calibrationonfacialexpressions}). 
%Furthermore, we can see a closer resemblance of the ground truth correlations in the trained model.
\\ 
%TODO binary crossentropy loss weiter oben erwaehnen? 
%TODO Falls das noch kommt: (We did add them to the related work section though because we understand why you regard them as similar.) 
%
% TODO Grafiken (Zahlen) größer machen, AU in plot einfügen
\section{Related Work}\label{relatedwork}

There are different ways to include domain knowledge in deep learning models as a constraint. Borghesi et al.~\cite{borghesi2020improving} highlight the following main approaches: feature engineering, modelling the hypothesis space e.g. with a Graph Neural Network~\cite{scarselli2008graph}, using constrained data augmentation, and adding a regularization term that includes mathematically formulated constraint knowledge. Based on these categories, we can categorize the related work with respect to our approach:
\paragraph{Regularization Term} A regularization term for enforcing domain constraints needs to be mathematically formulated in such a way that the term is differentiable and therefore suitable for updating the weights in a neural network. To the best of our knowledge we are the first to incorporate co-occurrence domain knowledge in a loss function. However, there are approaches that formulate regularization terms based on other domain knowledge. Muralidhar et al.~\cite{muralidhar2018incorporating} incorporate monotonicity and approximation constraints in the loss function for predicting the solubility of oxygen in water. Song et al.~\cite{song2021cc} use an additional channel correlation loss for image classification in order to constrain the relations between classes and channels. Since these are different constraints, we cannot build on their approaches.
\paragraph{Model Hypothesis Space} Most approaches using AU correlations model this information in a hypothesis space. Corneanu et al.\cite{corneanu2018deep} apply Structure Inference that is inspired by graphical models to exploit the correlation information and update the AU predictions accordingly. Before applying Structure Inference, they use patch learning and fuse the patches.
%propose to combine patch learning (including fusing the patches) with applying Structure Inference that is inspired by graphical models and update the AU predictions accordingly to their correlation. 
Li et al.\cite{li2019semantic} use a Gated Graph Neural Network, which is guided by a knowledge graph containing AU correlations. Like Corneanu et al.\cite{corneanu2018deep}, they combine it with patch learning.
%use a multi-scale patch learning in order to learn AU regions that are then processed in a Gated Graph Neural Network, which is guided by a knowledge graph containing AU correlations. 
Cui et al.~\cite{cui2020knowledge} use a Bayesian Network to model the correlation information as a weak supervision for backpropagation. Song et al.\cite{song2021hybrid} employ different Bayesian graph structures to capture different correlations for each facial expression. In contrast to our work they do not use the correlations to finetune or evaluate on facial expressions.
All these approaches demonstrate increasing performance on the same dataset, which is encouraging but in contrast to us do not evaluate cross-dataset to demonstrate the model's generalizability. Furthermore, modeling the information in a hypothesis space is an extra step our approach does not need, which gives us for example the opportunity to finetune easily on new data.
%
%Cui et al.~\cite{cui2020knowledge} encode correlation information of AUs and between AUs and emotions into a Bayesian Network. They use the knowledge model as a weak supervision in the backpropagation and demonstrate increasing performance when using the embedded correlation information for training. They only evaluate within datasets and do not employ cross-dataset evaluation.
%
\paragraph{Other} Wang et al.~\cite{wang2020dual} use the correlation information between AUs and emotions as a probability for generating pseudo AU expressions in a semi-supervised approach. They evaluate their approach cross-dataset with three datasets, but do not compare with and without correlation information. Also, since it is a semi-supervised approach it is not suitable for comparison. Zhao et al.~\cite{zhao2015joint} propose a patch-learning approach that takes region patches of co-occurring AUs into account and thus preserve the correlation of the dataset.
%\textcolor{orange}{In this work, we brush ideas of Few-Shot Learning\cite{wang2020generalizing} and Transfer Learning\cite{zhuang2020comprehensive}, but our main focus is integrating domain knowledge.} 
Shao et al.~\cite{shao2021jaa} reach state-of-the-art results by learning AU detection and face alignment together. We compare our results to their J$\hat{A}$A model.\\
All in all, none of these approaches use the AU correlation information directly as a constraint in the loss function and we can conclude that there is a lack of cross-dataset evaluation. To the best of our knowledge we are also the first to use co-occurrence information for calibration on facial expression.
%
%Furthermore, to the best of our knowledge we are the first to utilize correlation information in a supervised setting to demonstrate the benefit of using co-occurrence information for model robustness by performing an extensive cross-dataset evaluation. Moreover, to the best of our knowledge we are also the first to use co-occurrence information for calibration on facial expression.\\
%
%
\begin{table*}[h]
	\renewcommand{\arraystretch}{1.3}
	\caption{Overview of different dataset properties. \textit{not spec.} stands for not specified, \textit{seq.} for sequence labeled}
	\label{table:datasets}
	\centering
	\begin{tabular}{|l||r|r|l|p{45mm}|p{45mm}|}
		\hline
		Dataset & Subjects & Frames & Coding  & Affect & Recording Setting \\
		\hline
		\hline
		Actor Study & 21 & 146,663 & frame & posed \& natural (actors reproduce AUs, emotions are triggered) & lab \\
		\hline
		AffWild2 & 47  & 303,250 & frame &  natural & in-the-wild videos (varying quality)\\
		\hline
		BP4D & 41  & 366,955 & frame & natural (subjects were stimulated)  & lab \\
		\hline
		CK+ & 123   & 10,734 & seq. & posed videos (neutral face to peak) & lab (mostly gray scale) \\
		\hline
		EmotioNet (manual) & not spec. & 24,598 & frame & natural & in-the-wild (web images)\\
		\hline
		GFT & 96 & 172,800 & frame  & natural (social situation) & lab\\
		\hline
	\end{tabular}
\end{table*}
%
%\begin{figure}[!t]
%	\centering
%	\includegraphics[width=\columnwidth]{ckplus_gft_pearson.png}
%	\caption{Ground truth correlation matrix for the GFT (lower half) and CK+ (upper half) datasets.}
%	\label{fig:corr_gft_ckplus}
%\end{figure}
%
%
\section{Methods}\label{methods}
\subsection{Data Pre-Processing}\label{sec:datasetprep}
Table \ref{table:datasets} describes the different properties of our datasets.
%TODO include? The number of subjects is included, where it is given. AUs are coded either frame-wise or per sequence (seq.). The subjects' affect is posed or natural and the recordings took place in lab or in-the-wild settings. 
For training, we load the video frames or images in color and crop the faces with the OpenCV~\cite{opencv_library} DNN module for face detection, a state-of-the-art and open-source framework. We resize images to 224x224 pixels and normalize by zero-centering each color channel with respect to the ImageNet dataset used for pretraining.\\
AU datasets are often imbalanced, because AUs are not equally represented in facial expressions. Because AU detection is a multi-label problem, we apply a multi-label balancing optimizer~\cite{Rieger_2020} that operates on unique label sets. We use the following values: weighting parameter $\lambda$=$0.00001$, iterations$=$4000, and maximum occurrence of one image $n_u$=$6$. We use slight augmentations as recommended in~\cite{Rieger_2020}.\\
\subsection{Pearson Correlation Coefficient}
\begin{figure}[!t]
	\centering
	\includegraphics[width=6cm]{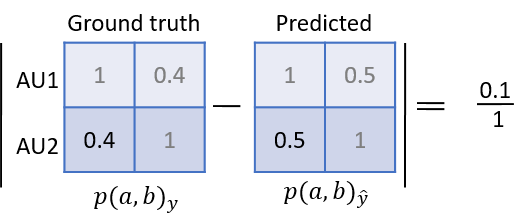}
	\caption{Visualization of the CorrLoss regularization term $c$ (see Eq. \ref{eq:corloss}). $p$ stands for pearson correlation.}
	\label{fig:corrloss}
\end{figure}
Because co-occurrence patterns between AUs vary greatly with dataset setting and task, we do not utilize literature AU correlation information, but take a dataset-based approach. For calculating the strength of co-occurrences between AUs in the datasets, we employ the pearson correlation coefficient. Eq.~\ref{eq:pearson} shows the pearson correlation $p(a,b)$ formula between two AU classes a and b, where $N$ is the batch size. The pearson correlation computes the ratio of the covariance of two variables, normalized by the product of their standard deviations. %For the loss calculation we compare two matrices, where one matrix is calculated with continuous predicted values of the model and one with binary ground truth labels, both in the range [0,1]. 
In the case of binary variables the pearson correlation returns the Phi coefficient, also known as Matthews correlation coefficient. 
%A strich ist nicht erklärt
\begin{equation}
	\label{eq:pearson}
	p(a,b) = \frac{\sum_{i=1}^{N}(a_i-\bar{a})(b_i-\bar{b})}{\sqrt{\sum_{i=1}^{N}(a_i-\bar{a})^2}\sqrt{\sum_{i=1}^{N}(b_i-\bar{b})^2}}
\end{equation}
The correlation coefficient ranges from $-1$ (strong negative correlation) to $1$ (strong positive correlation) with $0$ indicating that no relationship exists. Two classes are medium correlated when the value is in the range [0.3, 0.49]. %TODO könnte noch raus
%We use a non-weighted pearson formula for computational reasons which does not consider the number of occurrences per class. 
For our experiments, we select the following AUs, as they are the intersection of annotated AUs of our training datasets BP4D, GFT, and CK+: 1,2,4,6,7,10,12,14,15,17,23, and 24. Based on these AUs, we select for each of our training datasets the AUs that have at least a correlation strength of $0.4$ with one other AU. Figure~\ref{fig:corr_ckplus}, \ref{fig:corr_bp4d}, and \ref{fig:corr_gft} in the appendix show the ground truth correlation matrices of the datasets.\\
\subsection{Loss Function}\label{lossfunction}
For training our models, we apply for each batch the combined loss from the CorrLoss and BCE. The CorrLoss (Fig. \ref{fig:corrloss}) enforces a reduction of the distance between the ground truth and predicted co-occurrence matrix during training.\\ Eq.~\ref{eq:corloss} shows our CorrLoss $c$ for two AUs $a$ and $b$.  The variables $p(a,b)_{y}$ and $p(a,b)_{\hat{y}}$ note the pearson correlation for ground truth $y$ and predicted labels $\hat{y}$ respectively. Furthermore, we add 1 to make correlation values positive for computational reasons. In the denominator, the variable $U$ is the number of AU classes. The denominator scales the result by the number of correlations. %By the design of the CorrLoss, we optimize positive and negative AU correlations.
\begin{equation}
	\label{eq:corloss}
	c(a,b) = \frac{\bigl|\abs{p(a,b)_\text{y}  + 1} - \abs{p(a,b)_\text{\^{y}}  + 1}\bigr|}{0.5(U^2 - U)},\\ %for two classes 
\end{equation}
For reasons of simplicity it is omitted in Eq.~\ref{eq:corloss}, but when computing the CorrLoss for more than two AUs, the difference matrix is summed up. Also omitted is that we add a small error term $\epsilon$ to $y$ and $\hat{y}$ for numerical stability. Furthermore, to ignore double values in the matrix we apply a boolean matrix mask of the same size with values of 1 only above the diagonal. This mask would also offer the possibility to select specific correlations.
Regarding differentiability, CorrLoss is not differentiable when $p(a,b)_{y}$-$p(a,b)_{\hat{y}}$=$0$ because of the absolute value. But this is an unlikely event to occur for our experiments, since our averaged training loss is above 0 for every epoch. \\
%For derivation, only $p(a,b)_{\hat{y}}$ is dependent on the model's weights, all other variables are constant.\\
%TODO what impact has this? drin lassen?
Eq.~\ref{eq:binarycrossentropy} is the calculation of the BCE $e$, which we use since AU detection is a multi-class multi-label problem. $U$ is the number of AU classes.
%The variable $N$ is the batch size.
%
\begin{align}
	e = -\frac{1}{U}\sum_{i=1}^{U}(y_i\log\hat{y}_i+(1-y_i)log(1-\hat{y}_i)) 
	\label{eq:binarycrossentropy}
\end{align}
Eq.~\ref{eq:loss} is our final loss function $l$ combining the CorrLoss $c$ and BCE $e$. Dividing $c$ by two squishes the CorrLoss into the range [0,1] due to shifting the value range before. Variable $\rho$ weighs the CorrLoss and the BCE against each other. Setting $\rho$ to 1 ignores the BCE and setting it to 0 ignores the CorrLoss. We chose this design in order to be able to evaluate the influence of both terms in a grid search (Fig.~\ref{fig:gridsearch}).
%The crucial correlation weight $\rho$ is derived via grid search on the training dataset.
%
\begin{align}
	l = \frac{(1 - \rho)}{N}\sum_{i=1}^{N} e_i + \rho \frac{c}{2}
	\label{eq:loss}
\end{align} %binary crossentropy can be bigger as 1
\subsection{Evaluation Metric}
We use the macro F1-score (Eq.~\ref{eq:f1macro}) as our result metric because it compensates for both imbalanced classes and an imbalanced ratio of occurring and non-occurring AUs. Furthermore, applying this metric is widespread in the AU research field. Variable $tp$ stands for true positives, $fp$ refers to false positives, and $fn$ to false negatives.
\begin{align}
	F1_{macro} = \frac{1}{U} \sum_{i = 1}^{U} \frac{tp_i}{tp_i + \frac{1}{2}(fp_i + fn_i)}
	\label{eq:f1macro}
\end{align}
Our correlation metric $\bigl|\abs{p(a,b)_\text{y}  + 1} - \abs{p(a,b)_\text{\^{y}}  + 1}\bigr|$ shows how much the correlation matrix of the predicted values differs from the ground truth correlation matrix, where lower is better. We use it to measure the effect of our CorrLoss.
\subsection{Model and Training}
In order to facilitate comparability of our approach, we use a VGG16~\cite{simonyan2014very} neural network architecture, which is pretrained on ImageNet~\cite{deng2009imagenet}. Pretrained VGG16 Networks achieve good results for AU detection: Niinuma et al.~\cite{niinuma2019unmasking} evaluate different settings for AU detection learning such as pre-training and find out that generic pre-training using the ImageNet dataset yields better results than pretraining on face-specific datasets. We employ a 5-fold cross-validation with subject-dependent splits, which keeps subjects from leaking between training and validation data.
For training we freeze the weights up to layer \textit{block5\_conv3}. After the convolutional layers, we apply one fully connected layer with 256 neurons and a ReLu~\cite{nair2010rectified} activation function. A dropout layer~\cite{srivastava2014dropout} is placed between the fully connected layer and the output layer. A sigmoid activation function is used in the output layer. We save the best model when the validation loss is the lowest. For deterministic and reproducible training, we set seeds and use the deterministic implementation of TensorFlow when training on GPUs.
We use the Adam optimizer with a learning rate of $0.0001$ for all experiments. Gradient clipping with a clipvalue of $1$ proved to be beneficial. We train with a batch size of $64$ for all experiments.
\section{Results}
\subsection{Gridsearch}
\begin{figure}[t!]
	\centering
	\includegraphics[width=\columnwidth]{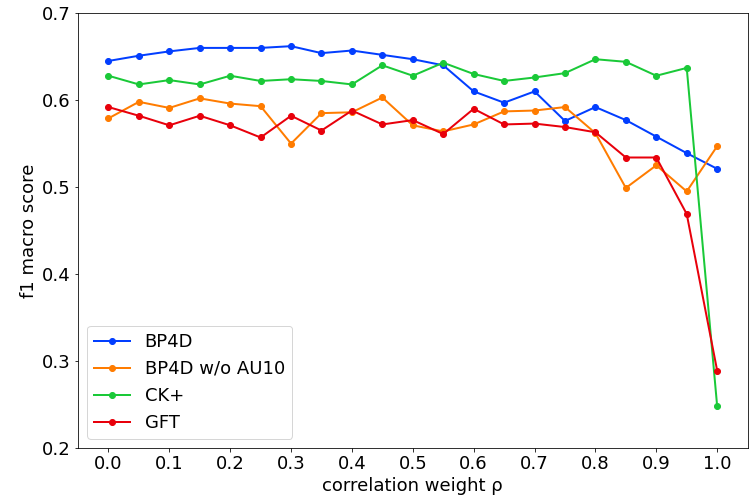}
	\caption{Gridsearch over different correlation weights $\rho$.}
	\label{fig:gridsearch}
\end{figure}

\begin{figure}[t!]
	\centering
	\includegraphics[width=\columnwidth]{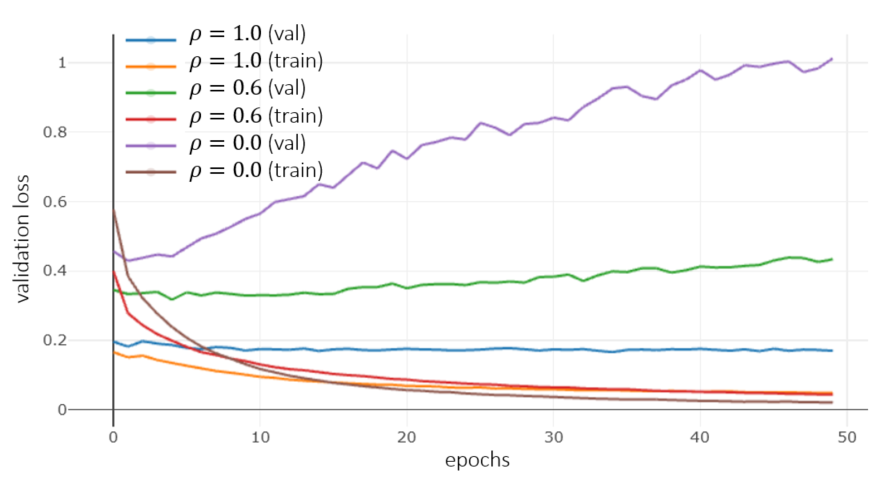}
	\caption{Training and validation loss of GFT gridsearch for different $\rho$.}
	\label{fig:gft_valloss_bc_fixfix}
\end{figure}

We use a gridsearch on all training datasets to determine the best correlation weight $\rho$ in each case for computing the CorrLoss. Figure~\ref{fig:gridsearch} shows the performance of the different $\rho$: We can see a difference between the frame labeled datasets BP4D and GFT and the sequence labeled dataset CK+. While BP4D with and without AU10 and GFT show a decreasing performance with a rising $\rho$, CK+ shows a high performance until a $\rho$ value of $0.95$. The CK+ dataset has more noisy frames since in each sequence the expression builds up from a neutral face to a maximum expression, but the whole sequence is labeled with the same AUs. Thus, it is possible that the CorrLoss can serve as a denoising measure.\\
We can see a lower but more consistent performance of the BP4D dataset over the different $\rho$ when training without AU10 (BP4D w/o AU10). The optimal $\rho$ is also higher ($0.3$ to $0.45$) when leaving the AU10 out, which is an indicator that our algorithm works better. The reason is that in BP4D only selected parts of sequences are labeled. It is therefore prone to an unusual imbalance: For some AUs, the number of occurrences is higher than the number of non-occurrences. AU10 occurs most often and therefore the neural network learns to predict the occurrence of AU10 above average. To not have this dataset bias impede our approach, we decide to train without AU10. We also conclude that the algorithm is sensitive to class occurrence distribution in the training set.
In general, our proposed loss function has a high performance for different $\rho$ and is therefore quite stable. For our training we choose the best $\rho$ per dataset.\\
Figure~\ref{fig:gft_valloss_bc_fixfix} shows the training and validation loss for the gridsearch. We can see that the overfitting decreases as $\rho$ increases. Thus, the CorrLoss likely serves as a regularization.
\begin{table*}[t!]
	\renewcommand{\arraystretch}{1.3}
	\caption{5-fold cross-validation. Results are in Macro F1-score. Lower variance is  underlined.}
	\label{table:kfold_bc_cv}
	\centering
	\begin{tabular}{|l||c|c||c|c|}
		\hline
		& \multicolumn{2}{c||}{not balanced}&  \multicolumn{2}{c|}{balanced} \\
		%\hline
		Dataset & $\rho$=0 & $\rho$$>$0 & $\rho$=0 & $\rho$$>$0\\
		\hline
		\hline
		CK+ ($\rho$=0.8) & 0.583 $\pm$ 0.027 & 0.604 $\pm$ \underline{0.026} & 0.592 $\pm$ 0.030 &  0.596 $\pm$ \underline{0.022} \\
		\hline
		%BP4D ($\rho$=0.3) & 0.617 $\pm$ 0.041& 0.613 $\pm$ 0.054& \textbf{0.623} $\pm$ \textbf{0.031} & 0.615 $\pm$0.032\\ % with AU14
		BP4D ($\rho$=0.3) & 0.615 $\pm$ 0.029 & 0.625 $\pm$ \underline{0.026}  & 0.636 $\pm$ 0.033& 0.629 $\pm$ \underline{0.025} \\
		\hline
		BP4D ($\rho$=0.45) & 0.587 $\pm$ 0.017 & 0.601  $\pm$ \underline{0.014} & 0.603 $\pm$ \underline{0.038} & 0.586 $\pm$ 0.039 \\
		\hline
		GFT ($\rho$=0.6) & 0.593 $\pm$ 0.020 & 0.594 $\pm$ \underline{0.015} &  0.616 $\pm$ \underline{0.017}& 0.613 $\pm$ 0.019 \\
		\hline
	\end{tabular}
\end{table*}
\subsection{Within Dataset Evaluation}
Table~\ref{table:kfold_bc_cv} gives an overview of our within dataset performance of the 5-fold cross-validation, which answers our first research question. The results show no clear tendency for better mean performance using the CorrLoss. %This meets our expectations since we constrain the Neural Network can develop in the training when using domain knowledge. 
%By using the constraint we set the focus on information less susceptible to noise in the training process. This does not stand out in the within dataset evaluation, since the validation dataset has the same pattern of noise. But it does in the cross-dataset evaluation.
%, tackling presumably spurious correlations and thus preventing overfitting on the training dataset. 
However, the variance of the 5-fold cross-validation decreases in most experiments using $\rho$$>$0. Since each fold contains different subjects, a smaller variance can mean a more stable and robust learning of AUs despite different appearances.
\subsection{Cross-Dataset Evaluation}\label{crossdatasetevaluation}
For our second research question we evaluate the cross-dataset performance (Table~\ref{table:cross_eval_bc_cv}). The higher mean result is in bold, if it is higher including variance, than the lower mean result. As earlier mentioned, each of our datasets display different properties regarding their recording conditions~(Table \ref{table:datasets}). Because of this, the results for cross-dataset cross-domain evaluation are generally much lower than for within dataset evaluation, which, however, is more realistic with regard to practical application~\cite{ertugrul2019}.
Our cross-dataset results show that for the majority of test datasets the mean performance is better and the variance is lower when applying the CorrLoss. By using the CorrLoss as constraint we may therefore set the focus on information less susceptible to noise in the training process. This is not apparent in the within dataset evaluation, because the validation dataset has the same noise pattern as the training dataset. However, it is in the cross-dataset evaluation. Thus, the cross-dataset results show the positive impact of the CorrLoss, highlighting the generalizing effect of our method to include correlation domain knowledge.\\
Looking at the bold mean results dataset-wise, the benefit of the CorrLoss is the highest for models trained on CK+. As CK+ is our smallest training dataset regarding number of frames and furthermore sequence labeled, this shows how domain knowledge can be beneficial, especially for small and noisy datasets. Furthermore, CK+ contains the most persons (123) of all datasets, which emphasizes that the CorrLoss helps in generalizing across subjects. % auch das einzige von den trainingsdatensets was geposed ist, noch in results ergänzen
The highest mean performance for the testing datasets is reached by models trained on the GFT dataset while using the balancing optimizer. This dataset contains the second most subjects (96), which emphasizes the generalization ability across subjects. Furthermore, GFT contains only one task, which might be beneficial.
As we already noticed in the gridsearch, BP4D is the most challenging training dataset. It includes a variety of different tasks and contains only 41 subjects.
%The correlation loss works well with the BP4D dataset when finetuning on different facial expressions (see Section~\ref{calibrationonfacialexpressions}).
%
Conclusively, variance needs to be considered when evaluating subject-dependent splits as each fold contains different subjects.
%Looking at the variance of the results is important because we employ a subject-dependent splitting and the different subjects in each fold contribute to the variance. 
As the variance is in most cases lower with the CorrLoss, it probably evens out the differences between subjects.
%Looking at the test datasets, the result on the AffWild2 dataset is overall the worst, but this is probably the most challenging dataset due to consisting of in-the-wild YouTube videos.
%Since we employ subject-dependent splits, there are different subjects in each fold, which explains the overall high variance. The constraint probably evens the differences between the subjects, since the co-occurrence of AUs stays comparably. 
%This supports the hypothesis that introducing co-occurrence information as a constraint has a positive effect on the robustness of the model evaluated on datasets with different properties.
% 
We observe that balancing~\cite{Rieger_2020} the multi-label training data is beneficial for the performance in most cases.
Although the performance in the cross-dataset evaluation is dataset-dependent, our results support the hypothesis that applying CorrLoss can have a positive effect on the generalizability and robustness of the models.
\begin{table*}[!t]
	\renewcommand{\arraystretch}{1.3}
	\caption{Cross-Dataset Evaluation. Results are in macro F1-score. Higher mean is bold and lower variance is underlined.}
	\label{table:cross_eval_bc_cv}
	\centering
	\begin{tabular}{|l|l||c|c||c|c|}
		\hline
		& & \multicolumn{2}{c||}{not balanced} & \multicolumn{2}{c|}{balanced}\\
		Train Dataset & Test Dataset & $\rho$=0 & $\rho$$>$0 & $\rho$=0 & $\rho$$>$0\\
		\hline
		\hline
		\multirow{5}{*}{\begin{tabular}{l}CK+: $\rho$=0.8\\(AU01, AU02, AU04, AU06, AU07, \\AU12, AU15, AU17, AU23, AU24)
		\end{tabular}} & Actor Study & 0.157 $\pm$ 0.009 & \textbf{0.183} $\pm$ \underline{0.005} & 0.179 $\pm$ 0.015& \textbf{0.197} $\pm$ \underline{0.010} \\
		\cline{2-6}
		&AffWild2 & 0.147 $\pm$  \underline{0.009} & \textbf{0.174} $\pm$ 0.012& 0.168 $\pm$ 0.009 & \textbf{0.182} $\pm$ \underline{0.005}\\ 
		\cline{2-6}
		&BP4D& 0.210 $\pm$ 0.015 & \textbf{0.245} $\pm$ \underline{0.008} & 0.247 $\pm$ 0.027 & \textbf{0.273} $\pm$ \underline{0.014} \\
		\cline{2-6}
		&EmotioNet Manual & 0.174 $\pm$ 0.015 & \textbf{0.218} $\pm$ \underline{0.014} &0.236 $\pm$ \underline{0.015} & 0.253 $\pm$ 0.019\\ 
		\cline{2-6}
		&GFT & 0.150 $\pm$ 0.016 & \textbf{0.203} $\pm$ \underline{0.008} & 0.191 $\pm$ 0.017 & \textbf{0.224} $\pm$ \underline{0.010}\\
		\hline
		\hline
		\multirow{5}{*}{\begin{tabular}{l}GFT: $\rho$=0.6\\(AU01, AU02, AU06, AU07, AU10, \\AU12, AU17, AU23, AU24)
		\end{tabular}} & Actor Study & 0.289 $\pm$ 0.013 & \textbf{0.301} $\pm$ \underline{0.011} & 0.304 $\pm$ \underline{0.006}& 0.309 $\pm$ 0.009 \\
		\cline{2-6}
		& AffWild2 &0.150 $\pm$ \underline{0.010}&  \textbf{0.169} $\pm$ 0.014& 0.199  $\pm$0.016 & \textbf{0.228} $\pm$ \underline{0.011}\\ 
		\cline{2-6}
		& BP4D& 0.431 $\pm$ \underline{0.020} & 0.416 $\pm$ 0.026 & 0.450 $\pm$ 0.034& \textbf{0.481} $\pm$ \underline{0.005} \\
		\cline{2-6}
		& CK+ & 0.303 $\pm$ 0.026 &  \textbf{0.324} $\pm$ \underline{0.019} & 0.336 $\pm$ 0.023& \textbf{0.359} $\pm$  \underline{0.014}\\
		\cline{2-6}
		& EmotioNet Manual & 0.355 $\pm$ 0.017 &  \textbf{0.375} $\pm$  \underline{0.016} & 0.419 $\pm$ 0.019& \textbf{0.438} $\pm$ \underline{0.007}\\ 
		\hline
		\hline
%		\multirow{5}{*}{\begin{tabular}{l}BP4D: $\rho$=0.3\\(AU01, AU02, AU06, AU07, AU10, \\AU12, AU17, AU24)
%		\end{tabular}} & Actor Study & \textbf{0.309} $\pm$ \textbf{0.008} & 0.294 $\pm$ 0.009& 0.291 $\pm$0.013& 0.293 $\pm$ 0.012\\
%		\cline{2-6}
%		&AffWild2 & 0.192 $\pm$ 0.015 & 0.186 $\pm$ 0.023& 0.198 $\pm$ \textbf{0.006} & \textbf{0.206} $\pm$ 0.009 \\
%		\cline{2-6}
%		&CK+ &0.353 $\pm$ 0.023 & 0.314 $\pm$ 0.029& 0.35 $\pm$ \textbf{0.015} & \textbf{0.362} $\pm$ 0.036\\
%		\cline{2-6}
%		&EmotioNet Manual & 0.310 $\pm$ 0.008 & 0.312 $\pm$ \textbf{0.006} & 0.328 $\pm$ 0.012 & \textbf{0.340} $\pm$ 0.015\\
%		\cline{2-6}
%		&GFT & \textbf{0.327} $\pm$ \textbf{0.010} & 0.323 $\pm$ 0.015 & 0.359 $\pm$ 0.015 & \textbf{0.363} $\pm$ 0.012\\
%		\hline
%		\hline
		\multirow{5}{*}{\begin{tabular}{l}BP4D: $\rho$=0.45 (w/o AU10)\\(AU01, AU02, AU06, AU07, AU12, \\AU17, AU24)
		\end{tabular}} & Actor Study &0.328 $\pm$ 0.017 & \textbf{0.342} $\pm$ \underline{0.013} & 0.316 $\pm$ 0.013 & 0.308 $\pm$ \underline{0.009} \\
		\cline{2-6}
		&AffWild2 & 0.180 $\pm$ 0.018 & \textbf{0.199} $\pm$ \underline{0.007} & 0.205 $\pm$ 0.006 & 0.206 $\pm$ \underline{0.005}\\
		\cline{2-6}
		&CK+ &0.373 $\pm$ \underline{0.026} & 0.401 $\pm$ 0.035 & 0.405 $\pm$ 0.022 & 0.422 $\pm$ \underline{0.019}\\
		\cline{2-6}
		&EmotioNet Manual & 0.234 $\pm$ 0.012 &  0.241 $\pm$ \underline{0.007} & 0.253 $\pm$ \underline{0.006} & 0.260 $\pm$ 0.013 \\
		\cline{2-6}
		&GFT & 0.320 $\pm$ \underline{0.012} & \textbf{0.340} $\pm$ 0.016 & 0.348 $\pm$ 0.013 & 0.350 $\pm$ \underline{0.010} \\
		\hline
	\end{tabular}
\end{table*}
\subsection{Calibration on Different Facial Expressions}\label{calibrationonfacialexpressions}
Facial expressions have individual co-occurrence patterns, where only a subset of AUs are involved. For our third research question we therefore explore the calibration of a trained model on facial expressions using our CorrLoss.
%This is the motivation for our third research question, where we explore to calibrate a trained model on specific co-occurrence patterns related to different facial expressions in order to gain better performance on unseen affective states.
We take the BP4D dataset, because it includes 8 different facial expression tasks for each person. We train 8 models, each without one task, where the model does not learn the target task, e.g. happiness. As a next step, we calibrate each model trained in this way on their target expression by finetuning it for 10 epochs on half of the frames. We use a subject-dependent split. For evaluating how good the model performs on the target expression after only finetuning, we test on the second unseen half of the task. Table~\ref{table:calibration_on_tasks_bp4d_feweraus} shows the testing results in macro F1-score on the left side and in correlation metric on the right side. We can see an overall performance increase when applying the CorrLoss with $\rho$=0.45. The correlation metric shows that the neural network trained only with BCE learns the co-occurrences automatically to a certain degree, but the model trained additionally with Corrloss captures the ground truth co-occurrences better.  
As we recall, Figure~\ref{fig:t1_pred} shows for the testing part of the task happiness visually that the learned correlations in a model trained with BCE and CorrLoss is closer to the ground truth correlations than when trained with only BCE. Quantitatively, Table~\ref{table:calibration_on_tasks_bp4d_feweraus} shows for this testing part a better correlation metric value of 6.6 when CorrLoss is used in contrast to 8.8 when only BCE is used.
% shows the ground truth correlation matrix (lower part) and predicted correlation matrix trained with our CorrLoss (upper half) of the testing part for the task happiness.
%
\begin{table}[!t]
	\renewcommand{\arraystretch}{1.3}
	\caption{Calibration on different facial tasks. Best results are bold.}
	\label{table:calibration_on_tasks_bp4d_feweraus}
	\centering
	\begin{tabular}{|l||c|c|c|c|}
		\hline
		& \multicolumn{2}{c|}{macro F1-score} & \multicolumn{2}{c|}{Corr. Metric}\\
		Tasks & $\rho$=0 & $\rho$=0.45& $\rho$=0 & $\rho$=0.45 \\
		\hline
		\hline
		1 - Happiness or Amusement  &0.61& \textbf{0.62} & 8.8& \textbf{6.6}\\
		\hline
		2 - Sadness &0.63& \textbf{0.70} &7.8& \textbf{6.4}\\
		\hline
		3 - Surprise or Startle& 0.55&  0.55 &11.8& \textbf{9.0}\\
		\hline
		4 - Embarrassment& \textbf{0.64}&0.61&14.0& \textbf{8.4}\\
		\hline
		5 - Fear or Nervous&0.61& \textbf{0.64}&15.2& \textbf{9.2}  \\
		\hline
		6 - Physical Pain&0.62& \textbf{0.68}&8.8&\textbf{7.6}\\
		\hline
		7 - Anger or Upset&0.49& \textbf{0.57}&9.8&\textbf{8.4}\\
		\hline
		8 - Disgust& \textbf{0.70 }& 0.68 &6.4& 6.4 \\
		\hline
	\end{tabular}
\end{table}
%
%\begin{figure}[!t]
%	\centering
%	\includegraphics[width=\columnwidth]{t1_pred_gt_cw045.png}
%	\caption{Ground truth correlation matrix (lower half) and predicted correlation matrix trained with CorrLoss (upper half) for testing part of task 1 (BP4D).}
%	\label{fig:t1_pred_gt} 
%\end{figure}
%
\subsection{ Comparison}\label{stateoftheartcomparison}
Table~\ref{tab:ckpluscomparison} shows a within dataset comparison of our model with several past and recent state-of-the-art approaches: JPML\cite{zhao2015joint}, DSIN\cite{corneanu2018deep}, SPERL\cite{li2019semantic}, AUD-EA\cite{cui2020knowledge}, J$\hat{A}$A~\cite{shao2021jaa}, and HMP-PS\cite{song2021hybrid}. Our model is trained with a 5-fold cross-validation using the balancing algorithm~\cite{Rieger_2020} and $\rho$=0.45. As the other approaches we apply a subject-dependent splitting. We compare the results for our 7 selected AUs. %Please note that the results for JPML\cite{zhao2015joint} stem from \cite{corneanu2018deep}. -> oder von hier? https://www.ri.cmu.edu/pub_files/2017/5/ant_low.pdf
%TODO?Furthermore, most papers claim to clean the training data in one or another way. While this is generally a good idea, lacking information about which data exactly was removed makes direct comparisons difficult.
Table~\ref{tab:ckpluscomparison} shows that our model reaches competitive performance on the BP4D dataset despite that in comparison with  other approaches our focus of optimization is not on increasing performance on the same dataset but on other unseen datasets. 
%Optimizing on the same dataset can lead to overfitting that is contrary to the generalization ability of a model. Hence, the performance of the model may be worse in a cross-dataset evaluation.
Table~\ref{tab:bp4dsota} shows a cross-dataset comparison from models trained on BP4D and evaluated cross-dataset on GFT. While J$\hat{A}$A~\cite{shao2021jaa} reaches great results on the within dataset evaluation, we significantly outperform in the cross-dataset evaluation, which means that our model has a greater generalization ability.
%TODO add that there are not other approaches with this cross-dataset approach
% search??? 
% https://www.mdpi.com/1424-8220/21/12/4222
% better still bp4d+ to gft: https://ieeexplore.ieee.org/stamp/stamp.jsp?arnumber=9667048&casa_token=waKhawZGWbMAAAAA:QBCgV923TlH7oLFWi5lBa47Bk3bhL3T1CD6sGambpptRWCA4o8B7GABlBUKSDEOFAxtPIk04Ggw1
% more state of the art results for BP4D https://openaccess.thecvf.com/content/CVPR2021/papers/Jacob_Facial_Action_Unit_Detection_With_Transformers_CVPR_2021_paper.pdf
% 
%
\begin{table}[!t]
	\renewcommand{\arraystretch}{1.3}
	\caption{Within dataset comparison with state-of-the-art approaches. Best results in macro F1-score are bold.} % balanced
	\label{tab:ckpluscomparison}
	\centering
	\begin{tabular}{|l||r|r|r|r|r|r|r|r|}
		\hline
		AU & 1 & 2 & 6 & 7 & 12 & 17 & 24 & Avg \\
		\hline
		\hline
		JPML  & 0.33 & \textbf{0.68} & 0.42 & 0.51 & 0.74 & 0.40 & 0.42 & 0.50 \\ %with region learning and correlation information, results from here: https://arxiv.org/pdf/1803.05873.pdf \cite{zhao2015joint}
		\hline
		DSIN & 0.52 & 0.42 & 0.77 & 0.74 & 0.87 & \textbf{0.67} & 0.47 & 0.64\\ %use corr information in the last step in a graph model \cite{corneanu2018deep}
		\hline
		SPERL & 0.47 & 0.45 & 0.77 & 0.78 & \textbf{0.88} & 0.64 & 0.53 & \textbf{0.65}\\ %Semantic relationships guided representation learning for facial action unit recognition. -> Knowledge Graph for AU correlation \cite{li2019semantic}
		\hline
		AUD  & 0.53 & 0.36 & 0.68 & \textbf{0.83} & 0.57 & 0.65 & - & 0.60 \\ %weakly supervised using correlation information \cite{cui2020knowledge} AUD-EA
		\hline
		J$\hat{A}$A & \textbf{0.54} & 0.48 & \textbf{0.79} & 0.76 & \textbf{0.88} & 0.62 & 0.50 & \textbf{0.65}\\
		%Z. Shao, Z. Liu, J. Cai, and L. Ma. Jˆaa-net: Joint facial action unitdetection and face alignment via adaptive attention. International Journal of Computer Vision, 2020.
		%b https://research.monash.edu/en/publications/j%C3%A2a-net-joint-facial-action-unit-detection-and-face-alignment-via
		\hline
		HMP & 0.53 & 0.46 & 0.77 & 0.77 & 0.86 & 0.63 & \textbf{0.55} & \textbf{0.65} \\ %\cite{song2021hybrid} hybrid message passing
		\hline
		ours & 0.51 & 0.41 & 0.74 & 0.69 & 0.84 & 0.53 & 0.31 & 0.58 \\ % AU_allcorr_newdf_bp4d_cv_fixfix_fewerAUs_bal 
		%\hline
		%ours & 0.43 & 0.26 & 0.76 & 0.79 & [0.86] & 0.57 & 0.29 & 0.57 \\ % AU_allcorr_newdf_bp4d_cv_fixfix_fewerAUs 
		\hline
	\end{tabular}
\end{table}
%ours AU_allcorr_newdf_bp4d_cv_fixfix_fewerAUs_bal 
%
%newest results for BP4D AU detection in https://ietresearch.onlinelibrary.wiley.com/doi/pdf/10.1049/ipr2.12480 
\begin{table}[!t]
	\renewcommand{\arraystretch}{1.3}
	\caption{Cross-dataset evaluation from BP4D to GFT. Best results in macro F1-score are bold.} % balanced
	\label{tab:bp4dsota}
	\centering
	\begin{tabular}{|l||r|r|r|r|r|r|r|r|}
		\hline
		AU & 1 & 2 & 6 & 7 & 12 & 17 & 24 & Avg \\
		\hline
		\hline
		J$\hat{A}$A & \textbf{0.14} & 0.19 & 0.34 & -& 0.15 & - & 0.02 & 0.17\\ %results are from this paper https://ieeexplore.ieee.org/stamp/stamp.jsp?arnumber=9667048&casa_token=waKhawZGWbMAAAAA:QBCgV923TlH7oLFWi5lBa47Bk3bhL3T1CD6sGambpptRWCA4o8B7GABlBUKSDEOFAxtPIk04Ggw1
		\hline
		%P-MCD & 0.10 & 0.22 & \textbf{0.54} & - & \textbf{0.57} & - & \textbf{0.32} & 0.35\\ %this is a domain adaption method which sees the target domain https://ieeexplore.ieee.org/stamp/stamp.jsp?arnumber=9667048&casa_token=waKhawZGWbMAAAAA:QBCgV923TlH7oLFWi5lBa47Bk3bhL3T1CD6sGambpptRWCA4o8B7GABlBUKSDEOFAxtPIk04Ggw1
		ours & 0.12 & \textbf{0.24} & \textbf{0.47}  & 0.53 & \textbf{0.46} & 0.42 & \textbf{0.26} &  \textbf{0.35}\\ %TODO achtung rundung!! 
		% Experiment: AU_allcorr_newdf_bp4d_cv_fixfix_fewerAUs_bal 
		%\hline
		%ours & 0.43 & 0.26 & 0.76 & 0.79 & [0.86] & 0.57 & 0.29 & 0.57 \\ % AU_allcorr_newdf_bp4d_cv_fixfix_fewerAUs 
		\hline
	\end{tabular}
\end{table}
%TODO in contrast to domain adaption methods we do not need to see the target images
%
\section{Conclusion \& Future Work}
%TODO add comparison with state of the art methods
We propose CorrLoss, a regularization term that formalizes domain knowledge about co-occurring facial movements as a constraint for training neural networks. We observe that applying CorrLoss improves the generalizability and robustness of neural networks when evaluated in a cross-dataset setting compared to our baseline. A comparison with state-of-the-art approaches confirms a higher performance in the cross-dataset evaluation and competitive but slightly lower performance in the within dataset evaluation. CorrLoss also proves to be effective against overfitting. Moreover, we show that CorrLoss can be applied to calibrate models on specific co-occurrences of different facial expressions.\\
Our CorrLoss could be used whenever predicted classes relate to each other--for example, when sub-concepts (e.g., eye) relate to higher-order concepts (e.g., human and animal)~(\cite{rabold2020expressive,finzel2021explanation}). This approach is therefore applicable to other domains as well.
%Thinking beyond images, this applies to all correlated information.
%
%Thinking beyond images, this also applies for time series data that is for example correlated to erroneous states.
%
% use Section* for acknowledgment
\section*{Acknowledgment}
We thank Robert Obermeier and Jan Adelhardt for helping in preprocessing the data. This work is funded by grant 01IS18056 A/B of BMBF ML-3 (TraMeExCo), 405630557 of DFG (PainFaceReader), and 16SV7945K of BMBF (ERIK).
\bibliographystyle{IEEEtran}
\bibliography{bib}
%% (used to reserve space for the reference number labels box)
%\begin{thebibliography}{1}
%\end{thebibliography}
\clearpage
\section*{Supplementary Material for \textit{CorrLoss: Integrating Co-Occurrence Domain Knowledge for Affect Recognition}}
\begin{figure}[h!]
	\centering
	\includegraphics[width=\columnwidth]{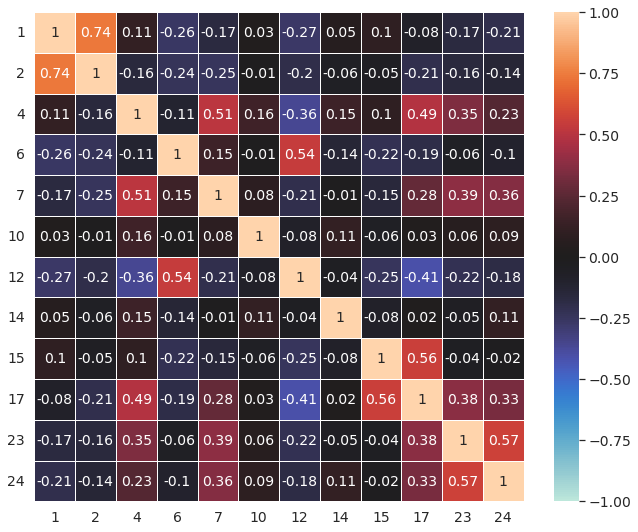}
	\caption{\textbf{CK+ dataset}: This is the ground truth correlation matrix. This dataset is sequence labeled and contains several tasks. Therefore the correlations are less distinct and more noisy.}
	\label{fig:corr_ckplus}
\end{figure}
\begin{figure}[h!]
	\centering
	\includegraphics[width=\columnwidth]{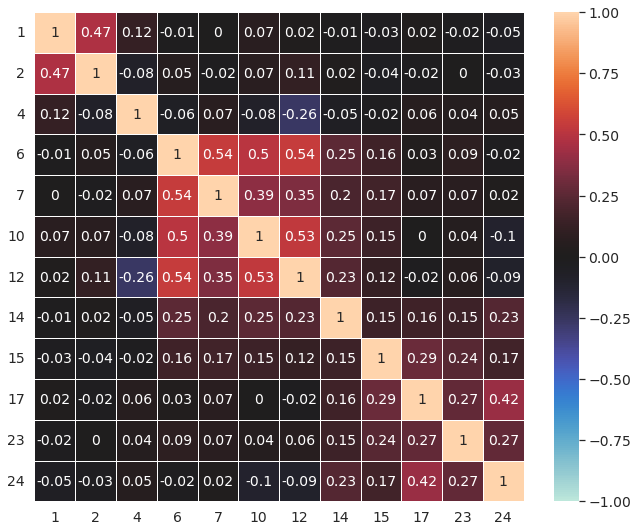}
	\caption{\textbf{BP4D dataset}: This is the ground truth correlation matrix.  This dataset contains several different facial expressions. It is visible that in most facial expressions eye brow movement (AU 1-2), cheek raising and lid tightening (AU 6-7), and mouth movements that extend also to the cheek and chin (AU 10-24), are present.}
	\label{fig:corr_bp4d}
\end{figure}
\begin{figure}[h!]
	\centering
	\includegraphics[width=\columnwidth]{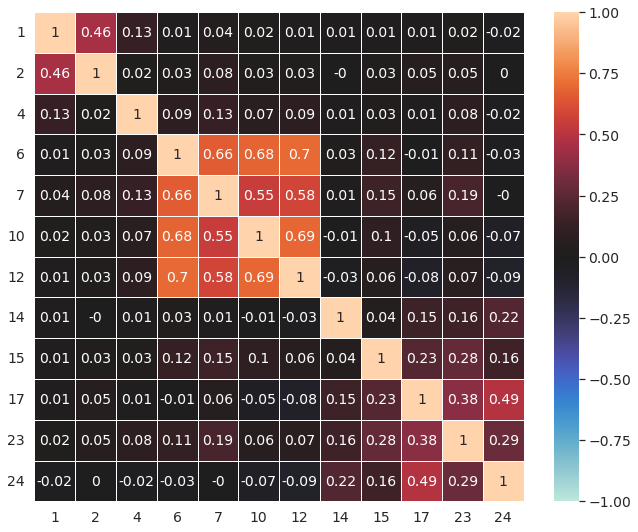}
	\caption{\textbf{GFT dataset}: This is the ground truth correlation matrix. We can see a clear pattern for this dataset because it incorporates only a single task in which the subjects are exposed to a social situation while drinking alcohol to make them relaxed. There is eye brow movement (AU 1-2), cheek raising and lid tightening (AU 6-7), and mouth movements that extend also to the cheek and chin (AU 10-24)}
	\label{fig:corr_gft}
\end{figure}
\end{document}